\documentclass[conference]{IEEEtran}
\usepackage{times}

\usepackage{fancyhdr}
\usepackage{xcolor} 
\usepackage{algorithm}
\usepackage{algorithmic}

\usepackage{microtype}
\usepackage{graphicx}
\usepackage{booktabs} 

\usepackage[numbers]{natbib}
\usepackage{multicol}
\usepackage[bookmarks=true]{hyperref}

\usepackage{amsmath}
\usepackage{amssymb}
\usepackage{mathtools}
\usepackage{amsthm}

\usepackage[capitalize,noabbrev]{cleveref}

\usepackage{booktabs, multirow}
\usepackage{amssymb}
\usepackage{pifont}
\usepackage{subfig}
\DeclareMathOperator*{\argmin}{argmin}

\definecolor{darkgreen}{rgb}{0,0.5,0}
\definecolor{darkyellow}{RGB}{153,153,0}
\definecolor{darkred}{rgb}{0.5,0,0}

\newcommand{\mcell}[1]{\begin{tabular}{@{}c@{}}#1\end{tabular}}

\newcommand{\mcelll}[1]{\begin{tabular}{@{}l@{}}#1\end{tabular}}

\theoremstyle{plain}
\newtheorem{theorem}{Theorem}[section]

\theoremstyle{definition}
\newtheorem{definition}[theorem]{Definition}

\theoremstyle{remark}

\begin{document}

\title{One-Shot Imitation Learning with Invariance Matching for Robotic Manipulation}

\author{Xinyu Zhang and  Abdeslam Boularias \\ 
Rutgers University \\ 
Email: \{xz653, ab1544\}@rutgers.edu }

\maketitle

\begin{abstract}
Learning a single universal policy that can perform a diverse set of manipulation tasks is a promising new direction in robotics. However, existing techniques are limited to learning policies that can only perform tasks that are encountered during training, and require a large number of demonstrations to learn new tasks. Humans, on the other hand, often can learn a new task from a single unannotated demonstration. In this work, we propose the Invariance-Matching One-shot Policy Learning (IMOP) algorithm. In contrast to the standard practice of learning the end-effector's pose directly, IMOP first learns invariant regions of the state space for a given task, and then computes the end-effector's pose through matching the invariant regions between demonstrations and test scenes. Trained on the 18 \textit{\textbf{RLBench}} tasks, IMOP achieves a success rate that outperforms the state-of-the-art consistently, by $4.5\%$ on average over the 18 tasks. More importantly, IMOP can learn a novel task from a single unannotated demonstration, and without any fine-tuning, and achieves an average success rate improvement of $11.5\%$ over the state-of-the-art on 22 novel tasks selected across nine categories. IMOP can generalize to new shapes and  objects that are different from those in the demonstration. Further, IMOP can 
perform one-shot sim-to-real transfer using a single real-robot demonstration. 

\let\thefootnote\relax\footnotetext{Project Website: \href{https://mlzxy.github.io/imop/}{\texttt{https://mlzxy.github.io/imop/}}. This work is supported by NSF
awards 1846043 and 2132972.}
\end{abstract}

\IEEEpeerreviewmaketitle

\section{Introduction}
\label{sec:intro}

Multi-Task Learning (MTL) is a recent type of learning technique where a single policy is trained to perform various tasks. With the recent advances in computer vision architectures~\citep{han2022survey}, a popular MTL strategy in robotics is to acquire a multi-task control policy through imitation learning from visual demonstrations~\citep{gervet2023act3d, goyal2023rvt, shridhar2023perceiver, xian2023chaineddiffuser}. These methods achieved impressive results in performing challenging manipulation tasks in unstructured 3D environments, such as RLBench~\citep{james2020rlbench}. However, these methods rely on the assumption that training and testing share the same set of tasks.
Moreover, training such policies on new tasks requires hundreds of demonstrations~\citep{goyal2023rvt}, which results in the catastrophic forgetting of old tasks~\citep{kirkpatrick2017overcoming}.


To address these issues, one-shot imitation learning aims to learn on a set of {\it base tasks} and generalize to {\it novel tasks} given only a single demonstration for each novel task and without re-training. However, existing methods rely on strong assumptions, such as requiring the novel tasks to be limited variations of the same base tasks~\citep{xu2022prompting}, requiring the base and novel tasks to have the same object setup~\citep{dasari2021transformers}, only generalizing specific actions on certain categories of objects with known 3D models~\citep{biza2023one}, or operating in simple 2D planar environments~\citep{duan2017one}. Moreover, most recent works focus on applying general popular techniques, e.g., transformers and diffusion models, to one-shot imitation settings, without taking advantage of the particular structure of robotic manipulation tasks~\citep{dasari2021transformers, mandi2022towards, xu2022prompting, xu2023hyper}. 

Therefore, a key question here is: can robots learn a manipulation policy that not only performs well on base tasks but also generalizes to novel unseen tasks using a single demonstration and without any fine-tuning? To this end, we propose IMOP (\textbf{I}nvariance \textbf{M}atching \textbf{O}ne-shot \textbf{P}olicy Learning), a new algorithm that not only outperforms the state-of-the-art on the standard 18 tasks of RLBench ($69.6\%$ average overall success rate compared to $65.1\%$ for the state-of-the-art~\citep{gervet2023act3d}), but also generalizes to 22 novel tasks with a single demonstration and without any fine-tuning ($41.3\%$ average success rate compared to $29.8\%$ for the state-of-the-art).  The 22 novel tasks are selected across nine categories and are substantially different from the base tasks in terms of objects and task goals~\citep{guhur2023instruction, james2020rlbench}. Moreover, we find that IMOP also generalizes to substantial shape variations, and can manipulate new objects that are different from those in the demonstration.

Instead of learning the desired end-effector's pose directly, IMOP learns key {\it invariant 
regions} of each task, and finds pairwise correspondences between the invariant regions in a one-shot demonstration and in a given test scene. The pairwise correspondences are used to analytically compute the desired end-effector's pose in the test scene from the least-squares solution of a point-set registration problem. 
The invariant region is defined as a set of 3D points whose coordinates remain invariant when viewed in the end-effector's frame, across states that share the same semantic action.
We devise a graph-based invariant region matching network. The invariant regions are located through neighbor attention~\citep{zhao2021point} from the KNN graphs that connect point clouds of demonstration and test scenes. The ground-truth invariant regions are discovered in offline unannotated demonstrations of base tasks.


To summarize, our contributions are threefold. (1) We propose IMOP, a one-shot imitation learning algorithm for robotic manipulation that learns a universal policy that is not only successful on base tasks, but also generalizes to novel tasks using a single unannotated demonstration. 
(2) We propose a correspondence-based pose regression method for manipulation tasks, which predicts robot actions by matching key visual elements, and a graph-based invariant region matching network on KNN graphs that connect demonstrations and test scenes. 
(3) We present a thorough empirical study of the performance and generalization ability of IMOP on a diverse set of tasks.

\section{Related Work}
\label{sec:related_work}

\textbf{Learning robotic manipulation from demonstrations.} Learning manipulation policies from offline visual demonstrations has gained increasing attention following the rapid improvement of vision models~\citep{dosovitskiy2020image,JunchiICRA2023,JunchiICRA2022,BoulariasKP11,BoulariasKP12,MuellingBoularias2014}. Prior works such as Transporter networks~\citep{zeng2021transporter} and CLIPort~\citep{shridhar2022cliport} learned simple pick-and-place tasks in a 2D planar setting. C2F-ARM~\citep{james2022coarse} and PerAct~\citep{shridhar2023perceiver} extend the control capacity to the 3D space with 6-DoF but learn a separate policy for each single task. More recently, RVT~\citep{goyal2023rvt}, Hiveformer~\citep{guhur2023instruction}, Act3D~\citep{gervet2023act3d}, and  ChainedDiffuser~\citep{xian2023chaineddiffuser} learned a multi-task policy from demonstrations of multiple tasks. Yet, these prior approaches can only perform tasks that were seen during training and cannot generalize to novel tasks upon inference. Training these models also requires hundreds of demonstrations per task. Instead, this work aims to develop a multi-task policy that not only performs well on seen tasks but also generalizes to novel tasks given a single demonstration.

\textbf{One-shot imitation learning. } Traditional imitation learning considers learning a policy for a single task with many expert trajectories provided for this particular task. 
One-shot imitation learning aims to learn from demonstrations of a set of base tasks and generalize to novel tasks using a single trajectory per new task without further training. \citet{duan2017one} trained a one-shot imitator that generalizes to simple task variations in a 2D stacking environment. \citet{finn2017one} used the same setting in the context of  meta-learning. \citet{dasari2021transformers, xu2022prompting} and \citet{mandi2022towards} trained transformer-based one-shot imitators, such as encoding expert trajectories as prompts. \citet{biza2023one} achieved one-shot generalization for grasping tasks where 3D models of the objects are available. \citet{xu2023hyper} trained adaptation layers through hyper-networks. Yet, none of these prior works considered 6D manipulation tasks in the context of one-shot learning. 
 Moreover, recent works focus on applying existing techniques, such as transformers and adaptation layers, to one-shot imitation settings, but did not investigate the inherent task structure of robotic manipulation. 
In contrast, this work proposes a one-shot imitation learner for 6D  manipulation by discovering invariant regions of each task.



\textbf{Invariance and affordance in manipulation. } Incorporating invariance into deep learning models has been shown to drastically increase data efficiency and generalization~\citep{brandstetter2021geometric, cohen2016group, cohen2018spherical, lyle2020benefits, van2020mdp}. \citet{graf2023learning} design image augmentations to learn viewpoint-invariant features for manipulation. Visual affordance is a region representation of action possibility~\citep{ardon2020affordances, mendonca2023structured}. For example, Where2Act learns a probability map for pushing and pulling at each pixel~\citep{mo2021where2act}. In this work, we propose the concept of \textit{invariant region}. Instead of camera viewpoint invariance, we train neural networks to predict regions whose positions remain invariant to the robot end-effector for a given task. Unlike affordance, the proposed invariant region is not used to represent action probability but to transfer actions from demonstrations to test scenes.



\section{One-shot Imitation with Invariance Matching}
\label{sec:method}

\begin{figure}
  \includegraphics[width=\linewidth]{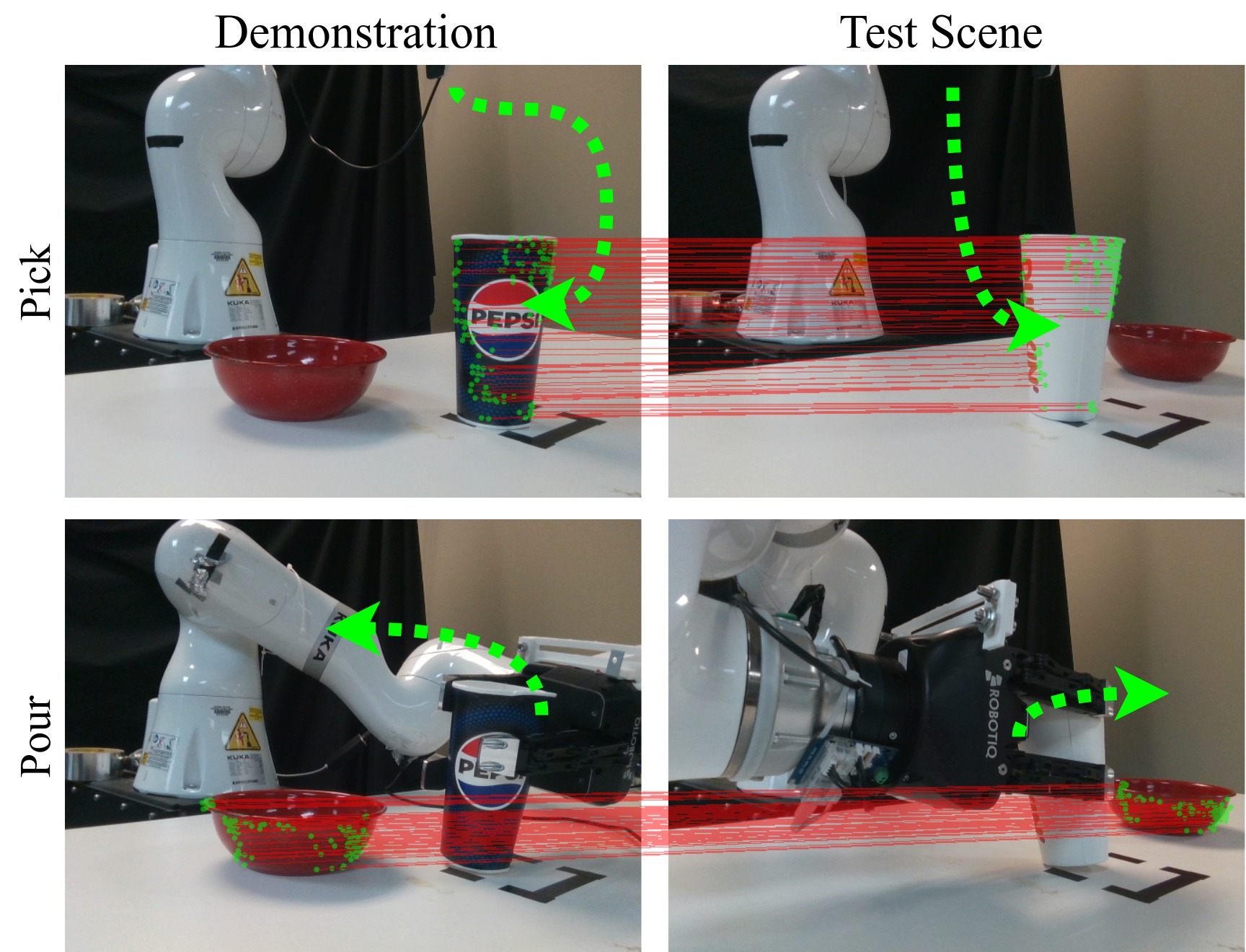}
  \caption{Example of a pick-and-pour task executed by a real Kuka robot after observing a single demonstration and without any re-training. 
  The correspondences between estimated invariant regions of the 3D point cloud in the demonstration and in the test scene are visualized in \textcolor{darkred}{red} lines. The invariant regions are predicted and matched by a neural network, trained offline.}
  \label{fig:motivating}
\end{figure}

\subsection{A Motivating Example}

Figure~\ref{fig:motivating} illustrates an example from our experiments where a Kuka robot is tasked with picking up a cup and pouring it into a bowl, using a single demonstration of picking up and pouring a different cup in a different position. 
At the core of IMOP lies the capacity to estimate and match invariant regions of the given task. By finding correspondences of invariant regions in the demonstration and in the test scene, the demonstrated actions can be transferred to the test scene. In this example, the invariant regions are: (1) a set of 3D points on the surfaces of the cups where contact with the fingertips occurs, and (2), a set of 3D points from the bowl's point cloud that capture its spherical concave shape. 
The invariant regions of the 3D point cloud are learned offline from unannotated demonstrations, on various objects and tasks.



\subsection{Formulation of One-shot Manipulation Learning Problem}
\label{sec:formulation}

\begin{figure*}
  \includegraphics[width=\textwidth]{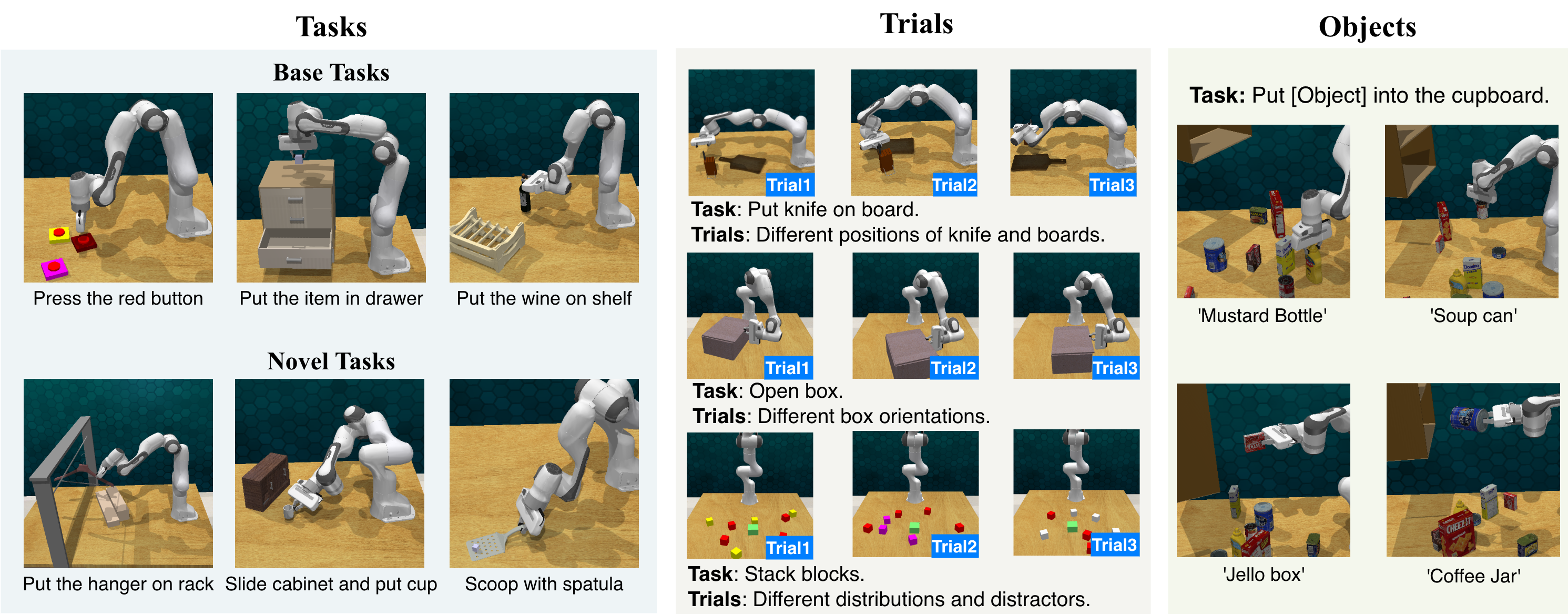}
  \caption{Examples of tasks, and object-level generalization in various trials. After training on base tasks, IMOP is evaluated on novel tasks that are substantially different from the base tasks. Every learned task is evaluated in multiple trials, each with different object layouts and orientations. For each novel task, only one recorded trajectory is given as a demonstration. We summarize the performance of IMOP on both base and novel tasks in Section~\ref{sec:exp-base} and \ref{sec:exp-novel}. Furthermore, we study the ability of IMOP to generalize to new objects beyond those in the demonstrations in Section~\ref{sec:exp-beyond-demo}.}
  \label{fig:rlbench}
\end{figure*}

We use $\mathbf{T}$ to denote a set of manipulation tasks, $\mathbf{T} = \{\textrm{Task}_1,\textrm{Task}_2,\dots,\textrm{Task}_{N_\textrm{tasks}}\}$. Each manipulation task $\textrm{Task}_i\in \mathbf{T}$ is performed in a Markov Decision Process (MDP) represented by 5-tuple $\mathcal{M}_i = (\mathcal{S}, \mathcal{A}, \mathbf{P}, R_i, \mu_i)$, wherein $\mathcal{S}$ and $\mathcal{A}$ are state and action spaces, $\mathbf{P}:\mathcal{S}\times\mathcal{A}\mapsto\mathcal{S}$ is a state transition probability, $R_i: \mathcal{S}\mapsto\mathbb{R}$ denotes a reward function associated with task $\textrm{Task}_i$, and $\mu_i$ is the initial state distribution of $\textrm{Task}_i$. $\mathbf{T}$ is composed of a set of base tasks, denoted by $\mathbf{T}^{base}$, and a set of novel tasks, denoted by $\mathbf{T}^{novel}$. Thus, $\mathbf{T} = \mathbf{T}^{base} \cup \mathbf{T}^{novel}$. 
During training, a large collection of offline demonstrations is provided for the base tasks. 
The underlying reward function of each task is unknown and is not used in this work.
During testing, only one successful trajectory $\tau_i = \{(s_i, a_i, s_i')\}^{|\tau_i|}$ is provided to the robot for a novel $\textrm{Task}_i\in \mathbf{T}^{novel}$, where $|\tau_i|$ denotes the number of transitions, $a_i$ is the demonstrated action in state $s_i$, and $s{_i}'$ is the resulting next state.
The goal is to learn a one-shot universal policy $\pi(s,\tau_i)$, which is a multi-task policy that returns an action $a$ in new state $s$ given a single demonstrated trajectory $\tau_i$ of the desired novel skill. $\tau_i$ is provided in an environment that is different from the one in which $\pi$ is tested. 




A state $s \in \mathcal{S}$ is a raw 3D point cloud of the entire scene, including the robot. Each point in state $s$ is represented in homogeneous coordinates. 
An action $a \in \mathcal{A}$ is the desired (i.e., target) 6D pose of the robot's end-effector, along with the desired state of the gripper. To execute an action, low-level robot movements are generated using off-the-shelf motion planners.
More precisely, we define actions as 18-dimensional vectors $a = (T, \lambda)$, wherein $T \in \mathbb{R}^{4\times4}$ denotes the target end-effector pose as a homogeneous matrix and is referred to as action pose, and $\lambda$ denotes two binary values that indicate whether gripper is open and collisions are allowed or not.


\begin{figure*}
  \includegraphics[width=\textwidth]{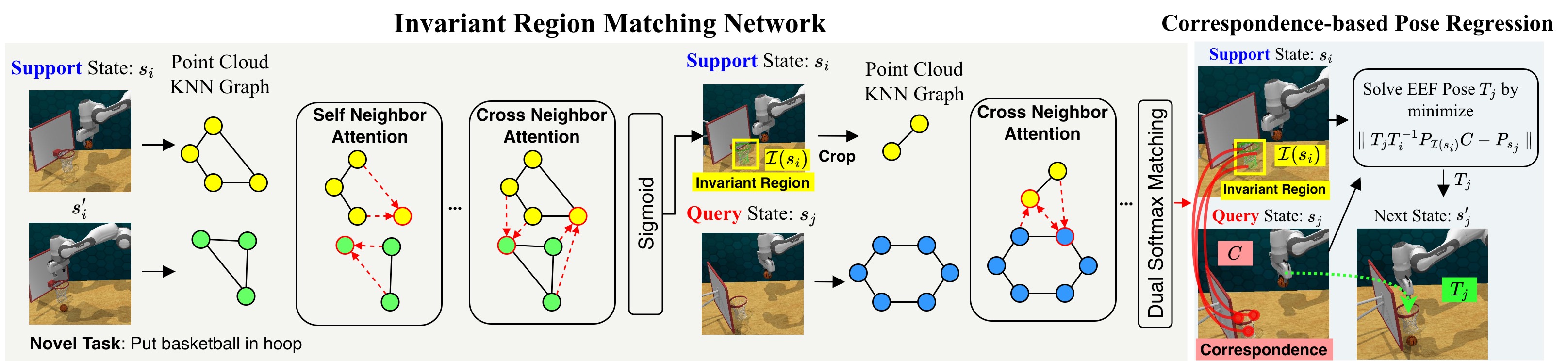}
  \caption{Overview of the proposed invariant region matching network and correspondence-based pose regression. Given the observed transition of the support state $s_i$, we build KNN graphs from the scene point cloud and apply graph self-attention (within $s_i$ and $s_i'$) and cross-attention (between $s_i$ and $s_i'$) layers. The invariant region $\mathcal{I}(s_i)$ is predicted as the set of activated points through a point-wise sigmoid over the support state $s_i$. Next, we apply graph cross-attention layers between the KNN graphs of $\mathcal{I}(s_i)$ and query state $s_j$. The correspondence matrix $C$ is predicted by matching the point-wise features $h_{\mathcal{I}(s_i)}$ and $h_{s_j}$. The action pose $T_j$ of query state $s_j$ is analytically solved from the action pose $T_i$ of support state $s_i$, correspondence matrix $C$, $P_{\mathcal{I}(s_i)}$ and $P_{s_j}$ (points in $\mathcal{I}(s_i)$ and $s_j$), as detailed in Section \ref{sec:corr-pose-reg}.
  }
  \label{fig:inv-reg-match}
\end{figure*}

\subsection{Proposed Method}

\subsubsection{Invariant Region Matching Network}We first consider the following problem. Given a task $\text{Task}_k\in \mathbf{T}$, an observed transition $(s_i, a_i, s_{i}')$ in a demonstrated trajectory, and a new state $s_j$ such that $s_i \equiv s_j$, how can action 
pose $T_i$ in state $s_i$ be translated to a new action pose 
$T_j$ to execute in the new state $s_j$? 
We use $s_i \equiv s_j$ to indicate that $s_i$ and $s_j$ share the same optimal manipulation action. For example, the first manipulation action of the task `open door' is `reach the door handle'. Therefore, states that describe different scenes with closed doors all share the optimal action `reach the door handle'. We refer to $s_i$ as the support state and to the new state $s_j$ as the query state. 

To answer this question, we define the notion of \textbf{invariant regions} in the following. 
\begin{definition}
    The invariant region of state $s_i$ in $\textrm{Task}_k$ is defined as $\mathcal{I}(s_i|\textrm{Task}_k) = \{ p \in s_i | \forall s_j \in \mathcal{S} \textrm{ s.t. } s_i \equiv s_j, 
\exists q \in s_j: \parallel T^{-1}_{i} p - T^{-1}_{j} q \parallel < \epsilon \}$, where $(T_i,\lambda_i) = \pi^*(s_i|\textrm{Task}_k)$ and $(T_j,\lambda_j) = \pi^*(s_j|\textrm{Task}_k)$. 
\label{def:inv}
\end{definition}
\noindent In this definition,  $s_i$ is a scene point-cloud, and \noindent $\mathcal{I}(s_i|\text{Task}_k)$ is the set of all the 3D points of $s_i$ whose coordinates in frame $T_{i}$ remain invariant across all the scenes $s_j$  that share the same optimal action as the optimal action of $s_i$. Frame $T_{i}$ is the frame of the end-effector's pre-contact pose when performing the optimal manipulation action in scene $s_i$. 
For example, for the task of picking up a cup, the optimal action is to grasp the cup's handle, regardless of the location, orientation, or style of the cup. The learned invariant region will likely be the cup's handle. 
To simplify the notation, we drop $\textrm{Task}_k$ and simply refer to the invariant region as $\mathcal{I}(s_i)$, while it is implied that $\mathcal{I}(s_i)$ is the invariant region of $s_i$ for a given $\textrm{Task}_k$. 

The optimal action pose $T_j$ of the query scene $s_j$ can be computed by first predicting the invariant regions $\mathcal{I}(s_i)$ and $\mathcal{I}(s_j)$ of $s_i$ and of support scene $s_j$, and then transforming $T_i$ according to a matching between $\mathcal{I}(s_i)$ and $\mathcal{I}(s_j)$. The overall framework of our proposed invariant region prediction and matching network is shown in Figure~\ref{fig:inv-reg-match}.  We first build a KNN graph for each scene point-cloud by connecting each point to its $k$ nearest points within the same scene. Next, we apply graph self-attention within each support scene $s_i$, and cross-attention between the KNN graphs of pairs of consecutive frames $s_i$ and $s_i'$ within the same support demonstration.
In contrast to traditional attention that is computed globally over all points, graph attention operates within the local neighborhood of each given point. We use the point transformer layer~\citep{wu2022point} as the graph attention operator. 
The invariant region $\mathcal{I}(s_i)$ is predicted as the set of activated points through a point-wise sigmoid over $s_i$. Then, we apply graph cross-attention layers between the KNN graph of $\mathcal{I}(s_i)$ and the query state $s_j$ to extract the point-wise features $h_{\mathcal{I}(s_i)} \in \mathbb{R}^{|\mathcal{I}(s_i)| \times D}$ and $h_{s_j} \in \mathbb{R}^{|s_j| \times D}$, where $D$ denotes the size of feature dimension. 
Finally, we perform a dual softmax matching~\citep{li2022lepard}   between $h_{\mathcal{I}(s_i)}$ and $h_{s_j}$ to obtain the correspondence matrix $C \in {[0, 1]}^{|\mathcal{I}(s_i)| \times |s_j|}$:

$$
C = \operatorname{softmax}(h_{\mathcal{I}(s_i)} \cdot h_{s_j}^\top) \cdot \operatorname{softmax}(h_{s_j} \cdot h_{\mathcal{I}(s_i)}^\top)^\top
$$

where $\operatorname{softmax}$ is applied on each row. The correspondence matrix $C$ maps each point in the invariant region $\mathcal{I}(s_i)$ to the query state $s_j$. The matched points in $s_j$ constitute $\mathcal{I}(s_j)$ but we only need $C$ to predict the action pose $T_j$ of the query state $s_j$, as explained in Section~\ref{sec:corr-pose-reg}. We detail the training of invariant region matching network in Section~\ref{sec:training}.



\vspace{0.5em}
\subsubsection{Correspondence-based Pose Regression}
\label{sec:corr-pose-reg}

The standard practice of 6-DoF pose regression is to obtain the  action pose $T$ from a neural network. However, this approach does not generalize well to new tasks, as shown in Section~\ref{sec:exp-novel}.
Instead, we propose to analytically compute action pose $T_j$ of the query state $s_j$ by solving the optimization problem in Equation~\ref{eq:solve} with a standard least-squares algorithm~\citep{huang1986least}, as follows,
\begin{equation}
T_j = \arg \min_{T\in \textrm{SE(3)}} \parallel T T^{-1}_{i}  P_{\mathcal{I}(s_i)}  C - P_{s_j} \parallel,
\label{eq:solve}
\end{equation}
\noindent where $T_i$ is the demonstrated action pose of the support 
state $s_i$,  $P_{\mathcal{I}(s_i)}$ and $P_{s_j}$ are the points in $\mathcal{I}(s_i)$ and $s_j$, $C$ is the predicted correspondence matrix. 
$C$ can be interpreted as an assignment matrix that maps each point in $\mathcal{I}(s_i)$ to a point in $s_j$.
Based on Definition~\ref{def:inv}, the optimal action pose $T_j$ is the solution for Equation~\ref{eq:solve} when the point mapping in correspondence matrix $C$ produces the minimal overall pairwise distance after applying  the transformation $T_j T_i^{-1}$.
Therefore, the pose regression problem can be solved by learning to match visual elements with Equation~\ref{eq:solve}. We use the differentiable Procrustes operator from Lepard~\citep{li2022lepard} to solve the least-squares problem.  As shown in Section~\ref{sec:exp-analysis}, our correspondence-based pose regression significantly improves the generalization performance.


It is worth noting that our correspondence-based pose regression shows a resemblance to the point cloud registration (PCR) problem~\citep{huang2021comprehensive}. However, there are three major differences. First, PCR assumes the point clouds are captured from the same scene with different camera viewpoints, while our method aligns point clouds from different scenes and objects. Second, PCR aligns the entire or the majority of two point clouds, while our method finds a sparser matching between only the invariant regions of some given task. Third, PCR finds the best matching, while our method uses the matching as an intermediate step to find the optimal pose of a robot's end-effector in a new scene.

\vspace{0.5em}
\subsubsection{State Routing Network}  
\begin{figure}
    \centering
    \includegraphics[width=\linewidth]{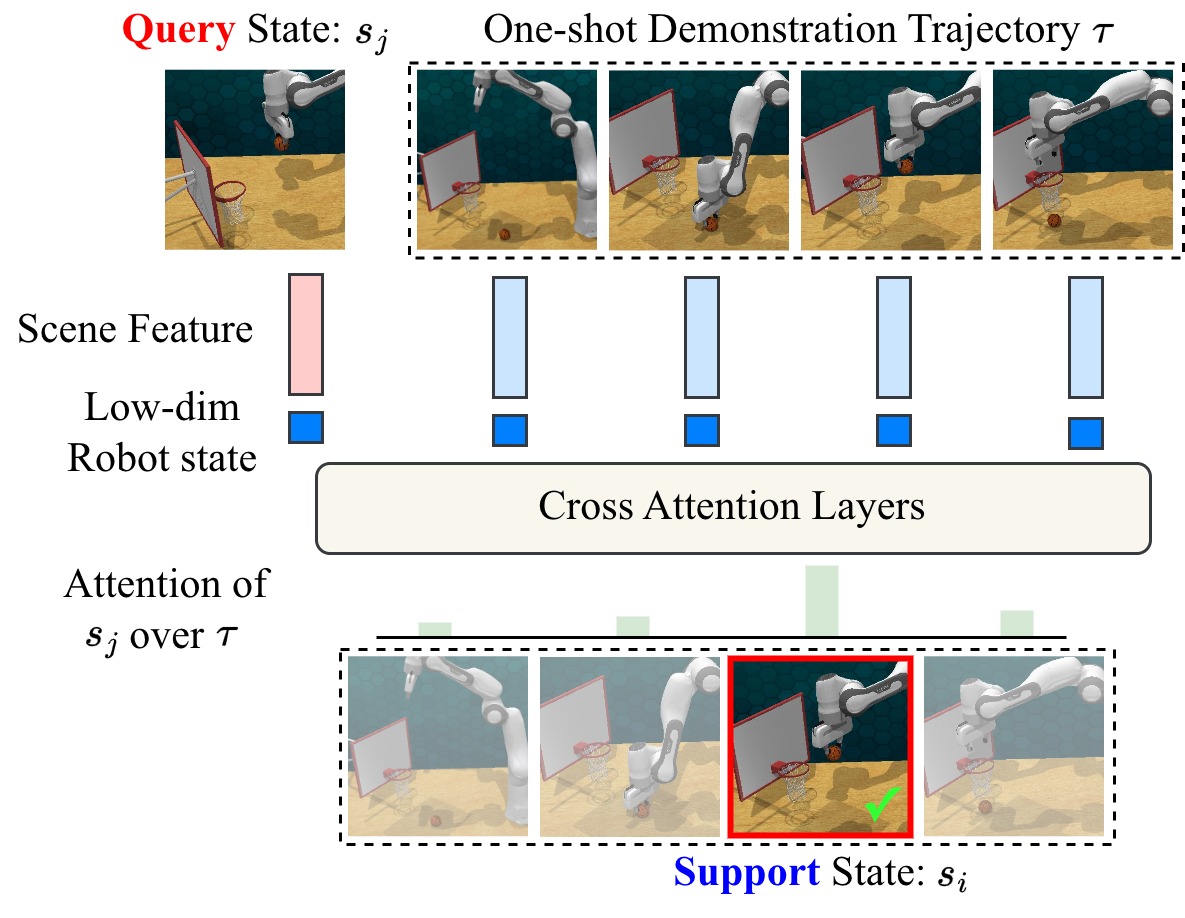}
    \vspace{-2em}
    \caption{Overview of the proposed state routing network. Given the query state $s_j$ and one-shot demonstration trajectory $\tau$, we extract the scene-level features for $s_j$ and each state in $\tau$. The scene features are concatenated with the corresponding low-dimensional robot states, and applied with cross-attention layers. The state with the strongest attention to the query state $s_j$ is selected as the support state $s_i$. }
    \label{fig:state-routing}
\end{figure}

We design the state routing network (shown in Figure~\ref{fig:state-routing})  to select the support frame $s_i$ in the one-shot demonstration $\tau$, given a query scene $s_j$.
We first extract the scene-level features for the query state $s_j$ and for each state in $\tau$ using a PTv2 backbone~\citep{pointcept2023}. Next, we follow the convention of existing work~\citep{goyal2023rvt} to concatenate the scene-level features with the low-dimensional internal robot state, including joint positions and timesteps. Then, we apply cross-attention over the features of multiple states. Unlike the invariant region matching network, classic attention layers are used instead of graph attention. Finally, we select the state with the strongest attention to the query state $s_j$ as the support state $s_i$.

The three techniques presented above form together the \textbf{I}nvariance \textbf{M}atching \textbf{O}ne-shot \textbf{P}olicy Learning (IMOP) algorithm. For each new query state $s_j$, the support state $s_i$ is determined from the one-shot demonstration trajectory $\tau$ by the state routing network. Next, the invariant region matching network predicts the correspondence matrix $C$ between $\mathcal{I}(s_i)$ and $s_j$. Finally, the action pose $T_j$ is derived from Equation~\ref{eq:solve}. We directly use the gripper values $\lambda_i$ of support state $s_i$ as the value of $\lambda_j$ for $s_j$.
By adapting the demonstration trajectory $\tau$, IMOP achieves one-shot generalization on novel tasks.

\vspace{0.5em}
\subsubsection{Training}
\label{sec:training}

To train IMOP, we assume the instance segmentation results are available. We use the segmentation masks provided by RLBench~\citep{james2020rlbench}. Note that 
IMOP does not require any segmentation during inference. 
Following the convention of RVT~\citep{goyal2023rvt} and PerAct~\citep{shridhar2023perceiver}, we represent each state as a macro-step using the keyframe extraction procedure from C2F-ARM~\citep{james2022coarse}. 
At each training iteration, two trajectories $\tau_i$ and $\tau_j$ are sampled if $|\tau_i| = |\tau_j|$. 
The state routing network is trained with focal loss~\citep{lin2017focal} to predict that $s_i \equiv s_j$ (i.e., $s_i$ and $s_j$ share the same manipulation action) if $s_i$ and $s_j$ have the same timestep and vice versa. Let $\{c_{i,n}\}^N_{n=1}$ denote $N$ instance segments, where each segment $c_{i,n} \in s_i$ is a set of 3D points of an object instance in $s_i$. To train the invariant region matching network, we estimate the ground-truth $\mathcal{I}(s_i)$ in Equation~\ref{eq:estimate-inv} as follows,
\begin{equation}
\mathcal{I}(s_i) = \arg \min_{c_{i, n}} \mathbf{E}_{\substack{p \in c_{i, n} \\ s_i \equiv s_j}} [ \min_{\substack{q \in c_{j, m} \\ \operatorname{cls}(c_{i,n}) = \\ \operatorname{cls}(c_{j,m})}} \parallel T_i^{-1} p - T_j^{-1} q \parallel ],
\label{eq:estimate-inv}
\end{equation}
\noindent where $\operatorname{cls}(c_{i,n})$ denotes the object class of the instance segment $c_{i,n}$. 
The ground-truth $\mathcal{I}(s_i)$ is the instance segment that has the minimal displacement in the frame of the action pose $T$. The ground-truth correspondence between $s_i$ and $s_j$ is the set of point mappings $\{(p, q) \in \mathcal{I}(s_i) \times \mathcal{I}(s_j)| q = \argmin_{q\in \mathcal{I}(s_j)} \parallel T^{-1}_{i} p - T^{-1}_{j} q \parallel\}$ such that the point-wise displacement is minimal in the action pose frame.

\begin{table*}[h]\centering
\caption{Base Task Performance on RLBench. We report the success rate for each task, and measure the average success rate and rank across all tasks. IMOP has the best overall success rate and rank. The success rate of IMOP is measured with an average of 5 runs. The success rates of baselines are reported from respective papers.}\label{tab:base}
\footnotesize
\begin{tabular}{lcccccccccc}\toprule
\multirow{2}{*}{Method} &Avg. &Avg. &Close &Drag &Insert &Meat off &Open &Place &Place &Push \\ 
&Success &Rank &Jar &Stick &Peg &Grill &Drawer &Cups &Wine &Buttons \\\midrule
C2F-ARM~\citep{james2022coarse} &20.1 &4.9 &24 &24 &4 &20 &20 &0 &8 &72 \\
PerAct~\citep{shridhar2023perceiver} &49.4 &3.6 &55.2 &89.6 &5.6 &70.4 &88 &2.4 &44.8 &92.8 \\
RVT~\citep{goyal2023rvt} &62.9 &2.3 &52 &\textbf{99.2} &11.2 &88 &71.2 &4 &91 &\textbf{100} \\
Act3D~\citep{gervet2023act3d} &65.1 &2.2 &\textbf{92} &92 &\textbf{27} &\textbf{94} &93 &3 &80 &99 \\
IMOP (Ours) &\textbf{69.6} &\textbf{1.9} &39.2 &98.4 &12.0 &92.0 &\textbf{100.0} &\textbf{52.8} &\textbf{96.0} &96.0 \\\midrule
&Put in &Put in &Put in &Screw &Slide &Sort &Stack &Stack &Sweep to &Turn \\
&Cupboard &Drawer &Safe & Bulb & Block & Shape & Blocks & Cups & Dustpan & Tap \\\midrule
C2F-ARM~\citep{james2022coarse} &0 &4 &12 &8 &16 &8 &0 &0 &0 &68 \\
PerAct~\citep{shridhar2023perceiver} &28 &51.2 &84 &17.6 &74 &16.8 &26.4 &2.4 &52 &88 \\
RVT~\citep{goyal2023rvt} &49.6 &88.0 &91.2 &48 &81.6 &36 &28.8 &26.4 &72 &93.6 \\
Act3D~\citep{gervet2023act3d} &\textbf{51} &90 &95 &47 &\textbf{93} &8 &12 &9 &92 &\textbf{94} \\
IMOP (Ours) &46.4 &\textbf{100.0} &\textbf{96.0} &\textbf{82.0} &58.4 &\textbf{37.6} &\textbf{40.0} &\textbf{56.8} &\textbf{100.0} &51.2 \\
\bottomrule
\end{tabular}
\end{table*}

\section{Experiments}
\label{sec:experiment}

We evaluate \textbf{IMOP} on the visual robotic manipulation tasks from RLBench~\citep{james2020rlbench}, a standard benchmark used in multi-task learning~\citep{james2022q}. We first train and evaluate our algorithm on the standard 18 RLBench tasks, and then measure its one-shot generalization ability on 22 novel tasks from 9 categories~\citep{guhur2023instruction}.
Further, we evaluate our method in real-robot experiments through one-shot sim-to-real transfer. 
Specifically, we aim to answer the following questions:

\begin{itemize}
    \item Can IMOP achieve high success rates on base tasks? 
    \item Can IMOP generalize to novel tasks given a single demonstration without tuning? 
    \item Can IMOP generalize to objects with large shape variations? 
    \item Can IMOP estimate meaningful invariant regions from demonstrations of novel tasks?
    
    \item Can IMOP be trained in simulation and transferred to solve real-robot manipulation tasks?
\end{itemize}

\textbf{Environment.} Following the convention~\citep{goyal2023rvt}, we record $128\times128$ RGB-D images from the front, left/right-shoulders, and wrist cameras. During training, we use 100 recorded trajectories per base task. 
RLBench contains various task categories, such as pick-and-place, tool use, high-precision operations, screwing, tasks with visual occlusion, and long-term manipulation. We choose 22 novel tasks that have different object setups and task goals from the base ones, according to the task categorization of Hiveformer~\citep{guhur2023instruction}. 
For each novel task, only a single successful trajectory is provided, as a one-shot demonstration.
Each task is evaluated on 25 independent trials, and we report the average success rate. RLBench generates different object layouts and orientations in different trials of the same task. For the base tasks, we use the same test scenes as the ones in~\citep{james2022q}. For novel tasks, we generate 25 test scenes for each task. Figure~\ref{fig:rlbench} illustrates the task,  and object-level generalizations of IMOP within the RLBench environment.
Further details are provided in the supplementary material.


\subsection{Performance Comparison on Base Tasks}
\label{sec:exp-base}

\textbf{Baselines.} To evaluate the performance of IMOP on base tasks, we test policies learned from offline demonstrations without the one-shot demonstrations. Specifically, we consider C2F-ARM~\citep{james2022coarse}, PerAct~\citep{shridhar2023perceiver}, RVT~\citep{goyal2023rvt}, and Act3D~\citep{gervet2023act3d}. C2F-ARM and PerAct learn separate policies for each task. Similar to our IMOP, RVT and Act3D learn a single multi-task policy. 

\textbf{Results.}  Table~\ref{tab:base} compares the success rates of IMOP and other baselines on the base tasks. C2F-ARM and PerAct use 3D voxel representation. RVT renders point clouds to virtual images. Similar to IMOP, Act3D uses raw point clouds as states. Overall, IMOP outperforms all baselines when the best rank and success rate are averaged across all tasks ($+4.5\%$ average success rate). IMOP has the highest rank on 10 out of 18 tasks. Remarkably, IMOP achieves a success rate of $52.8\%$ on the ``place cups" task while the existing works only reach a rate of $2\%$ to $4\%$. Moreover, not only it outperforms existing techniques, IMOP also generalizes to novel tasks, while the baselines can only be used on the tasks they were trained on.

\begin{figure*}
    \centering
    \includegraphics[width=\textwidth]{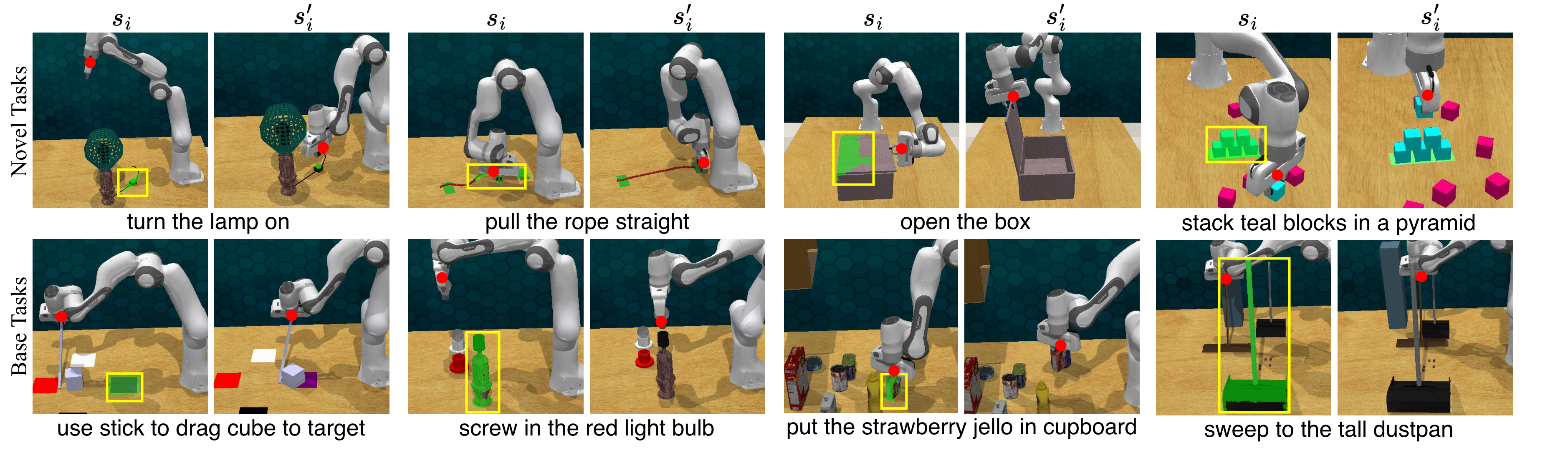}
    \caption{Visualization of invariant regions $\mathcal{I}(s_i)$ as estimated by our invariant region matching network on selected base and novel tasks. The invariant regions on $s_i$ are highlighted with \textcolor{darkgreen}{green} masks and \textcolor{darkyellow}{yellow} bounding boxes. The end-effector position is marked with \textcolor{darkred}{red} dot. $\mathcal{I}(s_i)$ generally covers the target object of the action that is applied by the robot to transition from state $s_i$ to next state $s'_{i}$. For example, the invariant region of `turn the lamp on' covers the `lamp switch' area. The invariant region of `pull the rope straight' covers one side of the rope and the rope target. It is worth noting that IMOP only takes point cloud as input and has no access to segmentation masks. }
    \label{fig:inv}
\end{figure*}

\subsection{Generalization to Novel Tasks}
\label{sec:exp-novel}

\begin{table*}[h]\centering
\caption{Novel Task Performance on RLBench. We report the success rate for each task, and measure the average success rate and rank across all tasks. IMOP has the best overall success rate and rank. Both RVT+FT and RVT+HDT require fine-tuning but IMOP does not. Directly using RVT without fine-tuning results in an average success rate of 0. We do not include this in the table. All success rates are measured with an average of 5 runs. }\label{tab:novel}
\footnotesize
\begin{tabular}{lcccccccccccc}\toprule
\multirow{2}{*}{Method} &Avg. &Avg. &Ball in &Rubbish &Scoop &Place &Hit &Block &Slide &Phone &Open &Close \\
&Success &Rank &Hoop &in Bin &Spatula & Hanger & Ball & Pyramid & Place &on Base & Box & Lid \\\midrule
RVT+FT &29.8 &2 &42.4 &50 &7.2 &4 &1.6 &0 &0 &20.4 &\textbf{28} &\textbf{100} \\
RVT+HDT &26.9 &2.4 &\textbf{87.6} &40.4 &0 &0 &0 &0 &0 &0 &24.8 &91.2 \\
IMOP (Ours) &\textbf{41.3} &\textbf{1.5} &80.0 &\textbf{94.0} &\textbf{52.0} &\textbf{24.4} &\textbf{32.0} &\textbf{5.2} &\textbf{6.8} &\textbf{22.8} &8.0 &71.2 \\\midrule
&Lid off &Lamp &Beat &Remove &Play &Knife on &Straighten &Change &Open &Open &Money  &Close \\
&Pan & On & Buzz &Cups &Jenga & Board &Rope & Clock &Bottle &Door &out Safe &Microwave \\\midrule
RVT+FT &78.8 &65.2 &19.2 &4 &40.8 &14 &0 &10.4 &34.4 &\textbf{68} &22.4 &44 \\
RVT+HDT &80.4 &\textbf{72} &16.4 &0 &27.6 &8 &0 &\textbf{36} &23.2 &12 &12 &\textbf{59.6} \\
IMOP (Ours) &\textbf{100.0} &12.0 &\textbf{20.0} &\textbf{36.0} &\textbf{100.0} &\textbf{20.0} &\textbf{16.8} &24.0 &\textbf{81.6} &41.2 &\textbf{42.8} &21.6 \\
\bottomrule
\end{tabular}
\end{table*}

\textbf{Baselines.} To assess the one-shot ability of IMOP, and since there is no existing one-shot learning method in the literature that was evaluated in RLBench, we extend the state-of-the-art technique RVT into two baselines, called RVT+FT and RVT+HDT, that can be used for one-shot learning. Specifically, since fine-tuning is the most common strategy for transferring models to new domains~\citep{yu2022survey}, we fine-tune RVT models pretrained on the base tasks with demonstrations of novel tasks and denote this baseline as RVT+FT. HDT~\citep{xu2023hyper} is a recent work that achieves one-shot policy generalization by training additional adaptation layers for new tasks while freezing original parameters. We implement the HDT adaptation layers in RVT and denote this baseline model as RVT+HDT. RVT and Act3D are both the most recent methods for learning multi-task robot control policies, and have similar performances, as shown in Table~\ref{tab:base}. We choose RVT over Act3D because the pretrained model of the latter is not yet available. For both RVT+FT and RVT+HDT, we train the respective network parameters with the one-shot demonstrations of novel tasks until convergence and adopt the sampling and 3D perturbation augmentation strategy from RVT and PerAct. 

\textbf{Results.} Table~\ref{tab:novel} compares the  success rates of IMOP, RVT+FT, and RVT+HDT on novel task. IMOP outperforms all baselines with the best rank and success rate averaged across all tasks ($+11.5\%$ over RVT+FT). IMOP has the best rank on 15 out of 22 tasks. Despite using the simple fine-tuning strategy, RVT+FT performs better than RVT+HDT on novel tasks. However, we observe the catastrophic forgetting phenomenon~\citep{kirkpatrick2017overcoming} in RVT+FT, whose average success rate on base tasks plummets to $4.6\%$ from $62.9\%$ after fine-tuning on novel tasks. RVT-HDT maintains the base task performance by freezing the original parameters but at the cost of having more parameters in the adaptation layers. Parameter sizes are compared in Table~\ref{tab:params}.

\begin{figure*}
\hspace{-1em}
  \includegraphics[width=\linewidth]{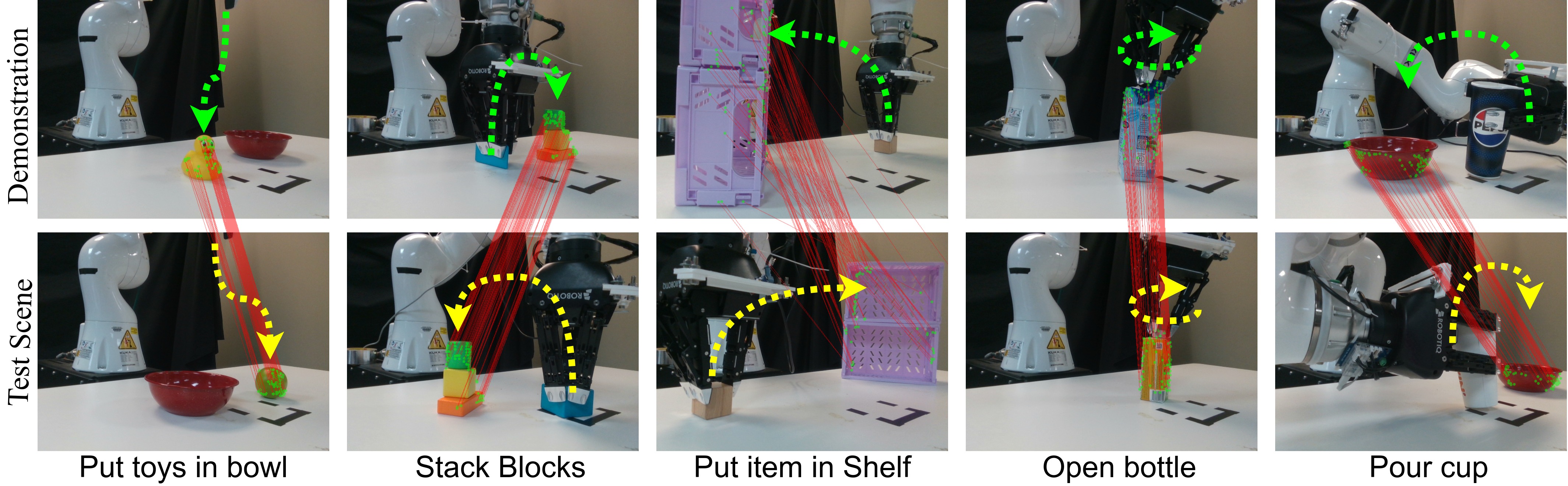}
  \caption{Real-robot manipulation examples of IMOP via one-shot sim-to-real transfer. The correspondences between the demonstration and test scenes are shown in \textcolor{darkred}{red} lines. The invariant regions closely overlap the next object to interact with. We observe that the one-shot generalization capacity of IMOP, trained with simulation data, enables the sim-to-real transfer using a single real-world demonstration without any re-training.}
  
  \label{fig:real-robot-exp}
\end{figure*}

\begin{table}[!htp]\centering
\caption{Network parameters size comparison between IMOP and one-shot baselines.}\label{tab:params}
\begin{tabular}{lcccc}\toprule
&\multicolumn{2}{c}{Success Rate (\%)} &\multirow{2}{*}{Params Size} \\
Method &Base &Novel & \\\midrule
RVT+FT &4.7 &29.8 &104M \\
RVT+HDT &62.9 &26.9 &131M \\
Ours &69.6 &41.3 &80.8M \\
\bottomrule
\end{tabular}
\end{table}

Compared to the baselines, IMOP generalizes to novel tasks by matching invariant regions between the one-shot demonstration and the states encountered during testing, without further training. Therefore, IMOP does not suffer from catastrophic forgetting or increased size of the parameters. Remarkably, both RVT+FT and RVT+HDT completely fail on certain tasks, \textit{e.g.}, ``block pyramid", ``slide place", and ``straighten rope", while IMOP still delivers a non-trivial performance on these tasks. Both ``block pyramid" and ``slide place" are categorized as long-horizon tasks and require over 30 and 10 macro-steps, respectively, to finish~\citep{guhur2023instruction}. The ``straighten rope" task requires operations on ropes, which are drastically different from the objects in the base tasks. This further indicates the limitation of the existing one-shot policy generalization strategies and the advantage of our proposed invariance matching-based policy. Figure~\ref{fig:inv} visualizes invariant regions on selected base and novel tasks. It can be seen that the invariant regions generally cover the target objects of the robot's actions. Additional visualizations are provided in the supplementary material.

\textbf{Analysis of underperformed tasks.} Despite the overall superiority of IMOP, the performance on specific tasks is still worth discussing. As in Table~\ref{tab:novel}, IMOP exhibits a lower success rate on `open box', `close microwave', and `close lid' compared to RVT+FT and RVT+HDT. These tasks involve manipulating flat surfaces, such as lids and box covers, as shown by the invariant region of `open box' visualized in Figure~\ref{fig:inv}. 
The challenge arises from the potential ambiguity in identifying invariant regions or correspondences on flat surfaces lacking rich geometric features. 
IMOP also performs worse than the baselines on the `lamp on' task, as you can see in the first two columns of Figure~\ref{fig:inv}. Although the invariant region estimation for this task is good, the area of the switch is small and hence, the correspondence is sparse compared to the whole scene, which leads to unstable solutions of the least-square problem in Equation~\ref{eq:solve}. The success rate of IMOP on long-term tasks, \textit{i.e.}, `block pyramid' and `slide place', outperforms the baseline methods but still has a large room for improvement. The IMOP mechanism of routing new states in test scenes to demonstrations is simple yet lacks the failure recovery ability because the demonstrations only contain successful trajectories. 
A more detailed analysis of failure cases is provided in the supplementary material.

\subsection{Generalization to Large Shape Variations}
\label{sec:exp-beyond-demo}

\begin{figure}
    \centering
    \captionsetup[subfigure]{justification=centering,labelformat=empty}
    \subfloat[`Spam' $\rightarrow$ `Crackers']{\includegraphics[width=0.49\linewidth]{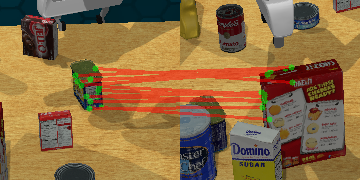}}\hfill
    \subfloat[][`Mustard' $\rightarrow$ `Sugar box']{\includegraphics[width=0.49\linewidth]{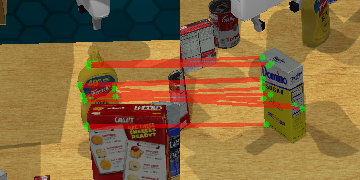}}

    \vspace{-0.8em}

    \subfloat[`Coffee jar' $\rightarrow$ `Soup can']{\includegraphics[width=0.49\linewidth]{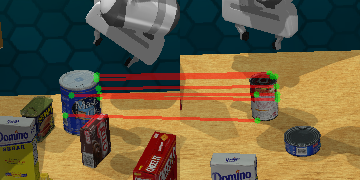}}\hfill
    \subfloat[][`Choco jello' $\rightarrow$ `Tuna can']{\includegraphics[width=0.49\linewidth]{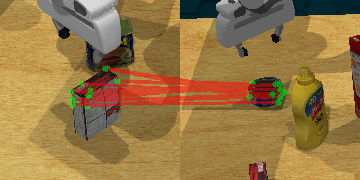}}
    \caption{Visualization of correspondence between different objects for the picking action. `Spam' $\rightarrow$ `Crackers' denotes using `Spam' as the demonstration to correspond to `Crackers' during testing.  
    Despite large shape variations, IMOP is able to estimate correspondences that effectively transfer actions, as shown by the success rate in Figure~\ref{fig:cross-object}.}
    \label{fig:corr-vis}
\end{figure}

\textbf{Setup.} Previous experiments evaluate the one-shot generalization ability of IMOP on novel tasks with one demonstration trajectory for each task. Even though there are large 
 variations in object shapes and categories between base and novel tasks, the variations within each novel task are limited.
For example, the novel task `put the knife on the chopping board' includes objects such as a knife, knife holder, and chopping board. 
None of these objects are seen during training and they are placed randomly at each trial and sometimes with color variations, but the object categories and shapes remain unchanged among various trials, as illustrated in Figure~\ref{fig:rlbench}. 
We are interested in whether IMOP can generalize to objects with large shape variations. 
To study this, we evaluate IMOP on objects different from those in demonstrations, which naturally have different shapes. Specifically, we consider a pick-and-place setting where the objective is to pick groceries and place them into a cupboard. 
The demonstration is given on a certain object, but picking and placing a different object is required during testing.
To simplify our study, we assume the access of the segmentation mask of the required object during testing.
Therefore, the correspondence between different objects can be obtained by masking out other regions.
To avoid memorization, we remove the corresponding state pairs from the training data.

\begin{figure}
  \includegraphics[width=\linewidth]{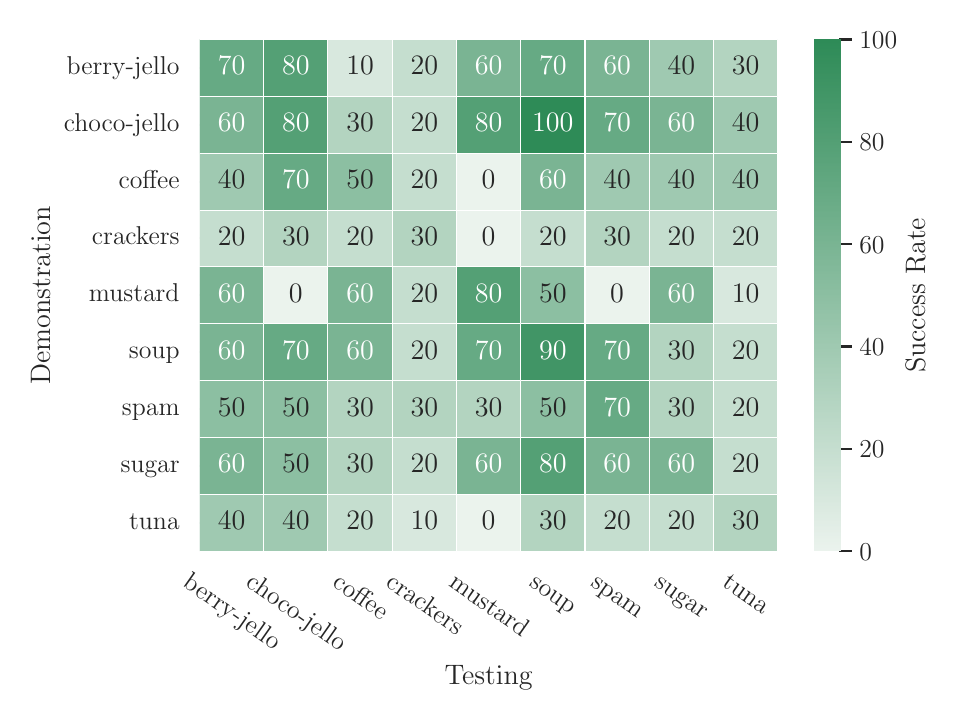}
  \caption{Success rate of pick-and-place task using different objects with large shape variations as demonstrations. For example, the entry at the top-right corner, \textit{i.e.}, `berry-jello, tuna', denotes the success rate is 30\% at picking and placing `tuna can' while using `berry-jello box' as the demonstration.}
  \label{fig:cross-object}
\end{figure}

\textbf{Result.} Figure~\ref{fig:cross-object} shows the success rate of pick-and-place with demonstrations on different objects. For example, the entry `mustard, coffee' represents the success rate of 
picking and placing a coffee jar with a demonstration on a mustard bottle.
There are 9 objects in total, and all success rates are measured on an average of 10 trials. It can be seen that IMOP performs competitively when a different object is used as a demonstration, except in some cases of the mustard bottle whose shape is drastically different than others. Using berry-jello and choco-jello as demonstrations delivers the same or even better results on many objects during testing, \textit{e.g.}, soup, spam, sugar, and tuna. These two jello boxes are the smallest objects in this setup.

Figure~\ref{fig:corr-vis} visualizes the correspondence between the invariant regions of different objects. The corresponding points, visualized as \textcolor{darkgreen}{green} dots, show a resemblance to the keypoint-based manipulation methods~\citep{gao2021kpam, manuelli2019kpam}. kPAM~\citep{manuelli2019kpam} and kPAM-SC~\citep{gao2021kpam} use object keypoints to compose objectives and constraints in the optimization-based planning for manipulation and demonstrate generalization over large shape variations in a pick-and-place setting. However, these methods require the manual design of keypoints for each object category. Our method IMOP does not require any specification of keypoints or any other form of labeling. The visualized connections are obtained by simply thresholding the correspondence matrix.

\subsection{Real-Robot Manipulation via One-shot Sim-to-Real Transfer}

\textbf{Setup.} We evaluate the one-shot sim-to-real performance of IMOP using simulation demonstrations for the base tasks, and trajectories recorded in the real-world as the one-shot demonstrations of novel tasks. We adopt five novel tasks: \textit{put toys in bowl}, \textit{stack blocks}, \textit{put item in shelf}, \textit{open bottle}, and \textit{pour cup}. Contrary to existing work that requires over 10 demonstrations per task for re-training~\citep{goyal2023rvt}, the one-shot generalization capability of IMOP enables the sim-to-real transfer using only one demonstration and without re-training. We deploy the model on a Kuka LBR iiwa robot. We use RBG-D observations from two RealSense D415 cameras. We use  MoveIt~\citep{moveit} for planning paths that move the arm to the gripper poses predicted by IMOP.

\begin{table}[!htp]\centering
\caption{One-shot sim-to-real transfer performance of IMOP for solving real-world manipulation tasks.} \label{tab:real-robot-exp}
\begin{tabular}{lcc}\toprule
\multirow{2}{*}{Task} &\multicolumn{2}{c}{Success} \\\cmidrule{2-3}
&One-shot RVT &IMOP (Ours) \\\midrule
Put toys in bowl & 3/10 &8/10 \\
Stack blocks & 1/10 &4/10 \\
Put item in shelf &2/10 &5/10 \\
Open bottle &2/10 &4/10 \\
Pour cup &2/10 &5/10 \\
\bottomrule
\end{tabular}
\end{table}

\textbf{Result.} Table~\ref{tab:real-robot-exp} shows the success rate of IMOP on the five real-world tasks. The success rate of each task is the average of independent 10 runs with different object layouts. Figure~\ref{fig:real-robot-exp} visualizes the correspondence between demonstration and test scenes. For the one-shot RVT, we finetune the pretrained RVT model with a single demonstration for each task. 
The sim-to-real gap is commonly acknowledged as a challenging problem in deploying control policies trained with simulation data~\citep{zhao2020sim}. 
In our study, we find that the region matching is robust to this domain shift. However, the sigmoid output of the invariant region prediction has much smaller values than those from simulation data. 
We circumvent this problem by thresholding the sigmoid output with a much smaller value, $0.1$.
This indicates that IMOP has the one-shot sim-to-real capacity, but still suffers from the sim-to-real gap for the invariant region estimation. This issue can potentially be addressed by fine-tuning with real-robot demonstrations or using more cameras to improve the point cloud quality.



\subsection{Ablation Studies}
\label{sec:exp-analysis}

\begin{table}[!htp]\centering
\caption{Ablation studies on the components of IMOP.}\label{tab:ablation}
\begin{tabular}{ll|cc}\toprule
\multicolumn{2}{c}{\multirow{2}{*}{\textbf{Configurations}}} &\multicolumn{2}{|c}{\textbf{Success Rate (\%)}} \\
& &\textbf{Base } &\textbf{Novel} \\\midrule
\multicolumn{4}{l}{\textit{State Routing Network}} \\\midrule
\multirow{2}[2]{*}{\mcell{Using Low-dim \\ robot states}} &Y &  69.6 &  41.3 \\\cmidrule{2-4}
&N & 63.2 & 33.0 \\\midrule
\multicolumn{4}{l}{\textit{Invariant Region Matching Network}} \\\midrule
\multirow{2}[2]{*}{Attention Type} &Graph Attention & 69.6 &  41.3 \\\cmidrule{2-4}
&Traditional Attention & 52.8 & 16.4 \\\midrule
\multirow{2}[1]{*}{\mcell{Matching \\ Strategy}} & \mcelll{Estimate $\mathcal{I}(s_i)$, then \\ match $\mathcal{I}(s_i)$ and $s_j$} & 69.6 &  41.3\\\cmidrule{2-4}
& Directly match $s_i$ and $s_j$ & 6.7 & - \\\midrule
\multirow{3}[3]{*}{KNN-Graph (K)} &4 &  58.7 &  32.4 \\\cmidrule{2-4}
&8 &  64.5 &  38.6 \\\cmidrule{2-4}
&16 & 69.6 &  41.3 \\\midrule
\multicolumn{4}{l}{\textit{Regression Method}} \\\midrule
\multicolumn{2}{l|}{Correspondence-based Regression (Equation~\ref{eq:solve})}  &  69.6 &  41.3 \\ 
\cmidrule{1-4}
\multicolumn{2}{l|}{Regression from matched points}  & 58.2 &  29.4 \\
\cmidrule{1-4}
\multicolumn{2}{l|}{Conventional pose regression}  & 65.1 &  - \\
\bottomrule
\end{tabular}
\end{table}

We study the component effects of IMOP in Table~\ref{tab:ablation}. For the state routing network, we find the low-dimensional robot state (i.e., joint positions and timesteps) contributes to the routing accuracy and success rate. This practice is also adopted by many existing work~\citep{shridhar2023perceiver, goyal2023rvt, gervet2023act3d}. 

For the invariant region matching network, we study the attention type, matching strategy, and the degree (K) of the point cloud KNN graphs. 
Table~\ref{tab:ablation} shows that KNN graphs of a larger degree provide stronger performance. However, using traditional attention instead of graph attention, which connects all points globally, significantly decreases performance. Note that we only apply traditional attention for region matching and still keep graph attention for invariant region estimation. This suggests local attention and propagating information through KNN graphs are effective inductive biases for 3D region matching. 
In contrast to the framework in Figure~\ref{fig:inv-reg-match}, which first estimates $\mathcal{I}(s_i)$ and then match $\mathcal{I}(s_i)$ and $s_j$, we investigate the strategy of directly matching $s_i$ and $s_j$ and predict the correspondence matrix. However, the model fails to predict actions on all tasks. 
This indicates the importance of estimating invariant regions. 

We compare our correspondence-based pose regression with two alternative solutions. The `regression from matched points' trains extra layers to predict pose from the matched points in correspondence matrix $C$ instead of deriving the action pose $T_j$ with Equation~\ref{eq:solve}. The `conventional pose regression' adopts a standard 3D regression head from Act3D~\citep{gervet2023act3d} while dropping the region matching mechanism completely. Table~\ref{tab:ablation} shows that conventional regression delivers comparable performance on base tasks but fails to generalize over novel tasks, and deriving action pose with Equation~\ref{eq:solve} performs better than predicting pose from matched points with an extra regression head.

\section{Discussion and Conclusion}

We have shown that one-shot novel task generalization can be achieved by learning to estimate and match key invariant regions in the demonstration and test scene. The target end-effector pose can be transferred by finding correspondences between invariant regions. Some open questions are, however, still worth discussing. 
IMOP is trained within the RLBench simulation environment, which offers segmentation masks and a validated key frame extraction procedure. 
For real-world data, a more sophisticated key frame extraction method and the use of open-set object detectors~\citep{zhang2023optical, zhang2023detect, zhong2022regionclip} may be necessary.
The proposed idea of transferring actions through matching key visual elements is general but the current definition of the invariant region is still closely linked to rigid body transformation. 
This suggests the potential for extending the formulation of invariant regions based on more general motion descriptors, such as warping or flows~\citep{li2022lepard}, to improve the manipulation accuracy on non-rigid objects and tasks with complex dynamics that cannot be regulated by poses. 
With the emergence of neural architectures that are scalable to Internet-scale data~\citep{brown2020language}, 
the possibility of learning invariant regions from unlabeled 2D videos that are widely available has yet to be fully explored. 
Beyond only leveraging a single demonstration, it is possible to maintain a pool of demonstrations and transfer actions from the most relevant state to improve the manipulation performance and reduce error accumulation under scenes with large variations or requiring failure recovery.




\end{document}